\documentclass[10pt,twocolumn,letterpaper]{article}

\usepackage{iccv}
\usepackage{times}
\usepackage{epsfig}
\usepackage{graphicx}
\usepackage{amsmath}
\usepackage{amssymb}

\usepackage{xcolor}
\usepackage{multirow}
\usepackage{adjustbox}
\usepackage{cancel}
\usepackage{soul}
\usepackage{footnote}
\usepackage[normalem]{ulem}
\usepackage{rotating}
\usepackage{subcaption}
\usepackage{color}
\usepackage{tabularx,ragged2e,caption}
\usepackage{wrapfig,lipsum,booktabs}
\usepackage{booktabs}

\newcommand{\fig}[1]{Fig.~\ref{#1}}
\newcommand{\tab}[1]{Tab.~\ref{#1}}

\usepackage[breaklinks=true,bookmarks=false,colorlinks]{hyperref}

\iccvfinalcopy % *** Uncomment this line for the final submission

\def\iccvPaperID{****} % *** Enter the ICCV Paper ID here

\ificcvfinal\fi

\begin{document}

%%%%%%%%% TITLE
\title{Active Learning for Deep Object Detection via Probabilistic Modeling}
%with Probabilistic Models
%(There is a problem that ICLR paper is searched if "mixture" is used in the title)

\author{Jiwoong Choi\textsuperscript{1,3},~Ismail Elezi\textsuperscript{2,3},~Hyuk-Jae Lee\textsuperscript{1},~Clement Farabet\textsuperscript{3},~and~Jose M. Alvarez\textsuperscript{3}\\
	\textsuperscript{1}Seoul National University,~~\textsuperscript{2}Technical University of Munich,~~\textsuperscript{3}NVIDIA\\
{\tt\small{\{jwchoi,~hjlee\}@capp.snu.ac.kr}},~~{\tt\small{ismail.elezi@tum.de}},~~{\tt\small{\{cfarabet,~josea\}@nvidia.com}}}

\maketitle
\ificcvfinal\thispagestyle{empty}\fi

%%%%%%%%% ABSTRACT
\begin{abstract}
%Active learning aims to reduce labeling costs by selecting only the informative samples to improve the network’s accuracy.
Active learning aims to reduce labeling costs by selecting only the most informative samples on a dataset. Few existing works have addressed active learning for object detection. Most of these methods are based on multiple models or are straightforward extensions of classification methods, hence estimate an image's informativeness using only the classification head. In this paper, we propose a novel deep active learning approach for object detection. Our approach relies on mixture density networks that estimate a probabilistic distribution for each localization and classification head's output. We explicitly estimate the aleatoric and epistemic uncertainty in a single forward pass of a single model. Our method uses a scoring function that aggregates these two types of uncertainties for both heads to obtain every image's informativeness score. We demonstrate the efficacy of our approach in PASCAL VOC and MS-COCO datasets. Our approach outperforms single-model based methods and performs on par with multi-model based methods at a fraction of the computing cost. Code is available at {\small\url{https://github.com/NVlabs/AL-MDN}}.

\end{abstract}

%%%%%%%%% BODY TEXT
\section{Introduction}
The performance of deep detection networks is dependent on the size of the labeled data \cite{liu2016ssd, ren2015faster}. Motivated by this, researchers have explored smart strategies to select the most informative samples in the dataset for labeling, known as active learning~\cite{DBLP:series/synthesis/2012Settles}. Typically, this is done by devising a scoring function that computes the network's uncertainty, selecting to label the samples for which the network is least confident with regard to its predictions \cite{DBLP:conf/cvpr/BeluchGNK18,Chitta2019Subsampling,DBLP:conf/cvpr/YooK19}.

\begin{figure}[!t]
	\centering
	\includegraphics[scale=0.67]{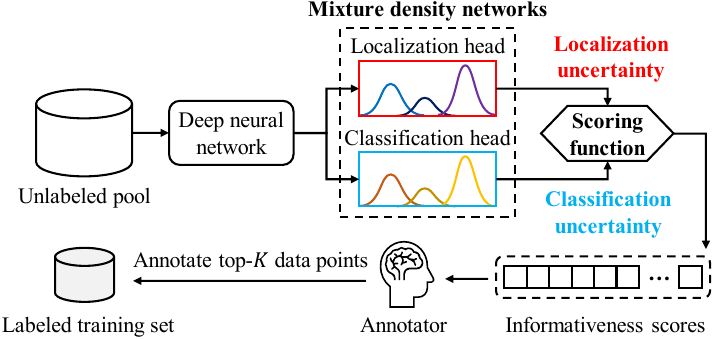}
	\caption{%A novel active learning approach for deep object detection with mixture density networks. 
	Our approach predicts aleatoric and epistemic uncertainties for both the localization and classification heads in a single forward pass of a single model. We propose an scoring function that aggregates epistemic and aleatoric uncertainties from both heads into  single value. Then, those data points with the top-\textit{K} scores are sent for labeling.} %and added to the labeled training set.}
	\label{fig:teaser}
    \vspace{-.3cm}	
\end{figure}

In general, the predictive uncertainty is decomposed into aleatoric and epistemic uncertainty~\cite{harakeh2020bayesod,HoraS:1996}. The former refers to the inherent noise in the data, such as sensor noise, and can be attributed to occlusions or lack of visual features~\cite{feng2018towards,NIPS2017_7141}. The latter refers to the uncertainty caused by the lack of knowledge of the model and is inversely proportional to the density of training data~\cite{NEURIPS2019_73c03186}. Modeling and distinguishing these two types of uncertainty is very important in active learning, as it allows the deep learning models to know about their limitations~\cite{hullermeier2019aleatoric,NEURIPS2019_73c03186}, \ie, recognize suspicious predictions in the sample (\textit{aleatoric uncertainty}) and recognize samples that do not resemble the training set (\textit{epistemic uncertainty}). To compute these types of uncertainty, researchers use multi-model based approaches, such as ensembles \cite{DBLP:conf/cvpr/BeluchGNK18} or Monte Carlo (MC) dropout~\cite{DBLP:conf/icml/GalIG17}. These methods reach good results but come with several limitations ~\cite{feng2019deep,haussmann2020scalable}. In particular, by being multi-models, they require a much higher computing cost, and in the case of ensembles, they also increase the number of the network's parameters ~\cite{DBLP:conf/cvpr/BeluchGNK18}. Additionally, they rely only on classification uncertainty, totally ignoring the localization uncertainty.%, despite being active learning for object detection. 

In this paper, we propose a novel active learning approach for deep object detection. Our approach uses a single model with a single forward pass, significantly reducing the computing cost compared to multiple model-based methods. Despite this, our method reaches high accuracy. To manage so, our method utilizes both localization and classification-based aleatoric and epistemic uncertainties. As shown in~\fig{fig:teaser}, we base our method on a mixture density networks~\cite{bishop1994mixture} that learns a Gaussian mixture model (GMM) for each of the network's outputs, \ie., localization and classification, to compute both aleatoric and epistemic uncertainties. To efficiently train the network, we propose a loss function that serves as a regularizer for inconsistent data, leading to more robust models. Our method estimates every image's informativeness score by aggregating all of the localization and classification-based uncertainties for every object contained in the image. We empirically show that leveraging both types of uncertainty coming from classification and localization heads is a critical factor for improving the accuracy. We demonstrate the benefits of our approach on PASCAL VOC~\cite{pascal-voc-2012} and MS-COCO~\cite{lin2014microsoft} in a single-stage architecture such as SSD~\cite{liu2016ssd}, and show generalized performance in a two-stage architecture such as Faster-RCNN~\cite{ren2015faster}. Our approach consistently outperforms single-model based methods, and compared to methods using multi-models, our approach yields a similar accuracy while significantly reducing the computing cost.

In summary, our \textbf{contributions} are the following:
\begin{itemize}
    \item We propose a novel deep active learning method for object detection that leverages the aleatoric and epistemic uncertainties, by considering both the localization and classification information. Our method is efficient and uses a single forward pass in a single model.    
    % \item We propose a novel loss to train the GMM-based classification head that leads to overall performance improvements in the network.
    \item We propose a novel loss to train the GMM-based object detection network that leads to overall performance improvements in the network.    
    %\item We propose a scoring function that leverages two types of uncertainties from the classification and localization heads in active learning. 
    %\item We propose an approach to reduce the modeling cost of the classification side of the mixture model.
    \item We demonstrate the effectiveness of our approach using different models on two different datasets.
\end{itemize}
\section{Related Work}
\textbf{Deep active learning for object detection} has recently acquired interest. 
The work of~\cite{haussmann2020scalable} trains an ensemble~\cite{DBLP:conf/cvpr/BeluchGNK18} of neural networks and then selects the samples with the highest score defined by some acquisition function, \ie, entropy~\cite{DBLP:journals/sigmobile/Shannon01} or mutual information~\cite{Chitta2018Large}. Concurrent work~\cite{feng2019deep} explores similar directions, but by approximating the uncertainty via MC-dropout~\cite{DBLP:conf/icml/GalG16,DBLP:conf/nips/KirschAG19}. The work of~\cite{DBLP:conf/iccv/AghdamGLW19} presents a method of calculating pixel scores and using them for selecting informative samples. Another approach~\cite{DBLP:conf/bmvc/RoyUN18} proposes a \textit{query by committee} paradigm to choose the set of images to be queried. The work of~\cite{DBLP:conf/iclr/SenerS18} uses the feature space to select representative samples in the dataset, reaching good performance in object detection \cite{DBLP:conf/cvpr/YooK19}. A different solution was given by~\cite{DBLP:conf/accv/KaoLS018} where the authors define two different scores: \textit{localization tightness} which is the overlapping ratio between the region proposal and the final prediction; and  \textit{localization stability} that is based on the variation of predicted object locations when input images are corrupted by noise. In all cases, images with the highest scores are chosen to be labeled. The state-of-the-art (SOTA) method of~\cite{DBLP:conf/cvpr/YooK19} offers a heuristic but elegant solution while outperforming the other single model-based methods. During the training, the method learns to predict the target loss for each sample. During the active learning stage, it chooses to label the samples with the highest predicted loss.

Most of the above-mentioned methods~\cite{feng2019deep,haussmann2020scalable,DBLP:conf/accv/KaoLS018} require multiple models or multiple forward passes to calculate the image's informativeness score, resulting in a high computational cost. In addition, all those studies, despite focusing in active learning for object detection, either rely on heuristic methods to estimate localization uncertainty \cite{DBLP:conf/accv/KaoLS018, DBLP:conf/cvpr/YooK19}, or cannot estimate it at all \cite{DBLP:conf/iccv/AghdamGLW19,feng2019deep,haussmann2020scalable,DBLP:conf/bmvc/RoyUN18,DBLP:conf/iclr/SenerS18}. Therefore, while giving promising directions, they are less than satisfactory in terms of accuracy and computing cost. In contrast to those methods, our approach estimates and leverages both the localization and classification uncertainties to reach high accuracy, while using a single forward pass of a single model, significantly reducing the computational cost.

\begin{figure*}[!t]
	\centering
	%\hspace{-0.0cm}
	\begin{tabular}{c|c}
	\includegraphics[scale=0.85]{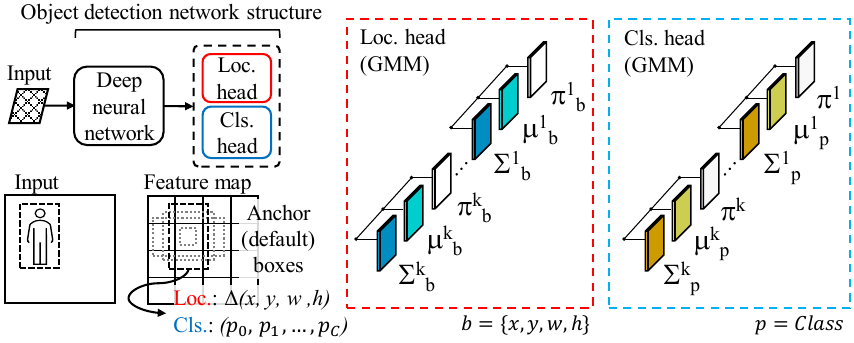}&
	\includegraphics[scale=0.85]{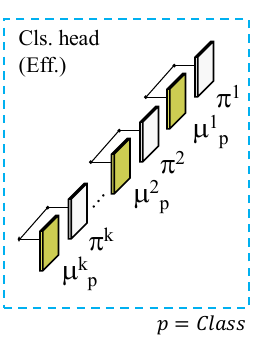}
    \vspace{-0.1cm}
	\\(a)&(b)
	\vspace{-0.25cm}
	\end{tabular}
	\caption{An overview of the proposed object detection network. The main difference with conventional object detectors~\cite{liu2016ssd,ren2015faster} is in the localization and classification heads (branches). a) Instead of having deterministic outputs, our approach learns the parameters of $K$-components GMM for each of the outputs: coordinates of the bounding box in the localization head and the class density distribution in the classification (confidence) head (see Section \ref{sec-main}). b) A classification head that improves the efficiency by eliminating variance parameters from GMM's classification head (see Section \ref{sec-efficient}).}
	\label{fig:network_re}
	%\vspace{-0.2cm}
\end{figure*}

\textbf{Mixture density networks} have been recently used for several deep learning tasks.
The approach of~\cite{choi2018uncertainty} focuses on the regression task for the steering angle. The works of~\cite{he2019deep, varamesh2020mixture} attempt to solve a multimodal regression task. The work of~\cite{yoo2019mixture} focuses on density estimation, while the work of ~\cite{choi2020task} attempts to explore the supervised learning problem with corrupted data. However, previous studies do not consider the classification task, which is an essential part of object detection ~\cite{choi2018uncertainty, he2019deep, varamesh2020mixture}. Additionally, all these studies do not take into account both types of uncertainty coming from the bounding box regression and the classification tasks ~\cite{choi2020task,choi2018uncertainty,he2019deep, varamesh2020mixture,yoo2019mixture}. Moreover, none of those studies address the problem of active learning for object detection. In contrast, our approach estimates and leverages both the aleatoric and epistemic uncertainties for both tasks in the context of active learning for object detection.
\section{Active Learning for Object Detection}
\label{headings}
The key novelty of our approach is designing the output layers of the neural network to predict a probability distribution, instead of predicting a single value for each output of the network (see \fig{fig:network_re}a). To this end, we propose to make use of a mixture density network where the output of the network consists of the parameters of a GMM: the mean $\mu^k$, the variance $\Sigma^k$, and the mixture weight $\pi^k$ for the \textit{k}-th component of the GMM. Given these parameters, we can estimate the aleatoric $u_{al}$ and epistemic $u_{ep}$ uncertainties~\cite{choi2018uncertainty}:
\begin{equation}\small
\begin{aligned}
u_{al}=\sum_{k=1}^K\pi^{k}\Sigma^{k},~~u_{ep}=\sum_{k=1}^K\pi^{k}\Vert\mu^{k}-\sum_{i=1}^K\pi^{i}\mu^{i}\Vert^2,
\end{aligned}
\end{equation}
where $K$ is the number of components in the GMM.
%Below we first introduce the mixture modeling for object detection for both localization and classification and then, we describe the scoring function to be used during active learning.

\subsection{Object detection with probabilistic modeling}
\label{sec-main}
To introduce our approach, we first focus on the localization task and then extend it to the classification task. As we will show later in our experiments, our method is applicable to both single-stage and two-stage object detectors.

\textbf{Localization:} In object detection, a bounding box $b$ is defined by its coordinates for the center ($x$ and $y$), its width ($w$), and its height ($h$). In our work, instead of predicting a deterministic value, our mixture model predicts $3$ groups of parameters for each bounding box: the mean ($\hat{\mu}_{x}$, $\hat{\mu}_{y}$, $\hat{\mu}_{w}$, and $\hat{\mu}_{h}$), the variance ($\hat{\Sigma}_{x}$, $\hat{\Sigma}_{y}$, $\hat{\Sigma}_{w}$, and $\hat{\Sigma}_{h}$), and the weights of the mixture ($\hat{\pi}_{x}$, $\hat{\pi}_{y}$, $\hat{\pi}_{w}$, and $\hat{\pi}_{h}$).

Let \{$\hat{\pi}^k_b$, $\hat{\mu}^k_b$, $\hat{\Sigma}^k_b$\}$_{k=1}^K$, $b\in\{{x, y, w, h}\}$ be the bounding box outputs obtained using our network. The parameters of a GMM with $K$ models for each coordinate of the bounding box are obtained as follows: 
\begin{equation}%\small
\label{eq:GMMpreproc}
\begin{aligned}
\pi^k_{b}=\frac{e^{\hat{\pi}^k_{b}}}{\sum_{j=1}^K e^{\hat{\pi}^j_{b}}},
~~\mu^k_{b}=\hat{\mu}^k_{b},~~\Sigma^k_{b}=\sigma(\hat{\Sigma}^k_{b}),
\end{aligned}
%\vspace{-0.8cm}
\end{equation}
\noindent where $\pi$ is the mixture weight for each component, $\mu$ is the predicted value for each bounding box coordinate, and $\Sigma$ is the variance for each coordinate representing its aleatoric uncertainty. As suggested in~\cite{choi2018uncertainty}, we use a softmax function to keep $\pi$ in probability space and use a sigmoid function to satisfy the positiveness constraint of the variance, $\Sigma_b^k>=0$.

\textbf{Localization loss: }The conventional bounding box regression loss, the smooth L1 loss~\cite{girshick2015fast}, only considers the coordinates of the predicted bounding box and ground-truth (GT) box. Therefore, it cannot take into account the ambiguity (\textit{aleatoric uncertainty}) of the bounding box. For training the mixture density network for localization, 
we propose a localization loss based on the negative log-likelihood loss. Our loss regresses the parameters of the GMM to the offsets of the center ($x$, $y$), width ($w$), and height ($h$) of the anchor (default) box ($d$) for positive matches:

\vspace*{-\abovedisplayskip}
\begin{equation}\small
\label{eq:localization_loss}
\centering
\resizebox{.99\hsize}{!}{$
\begin{gathered}
L_{loc}(\lambda,l,g)=-\sum_{i\in{Pos}}^N\sum_{b}\lambda^{ij}_Glog(\sum_{k=1}^{K}\pi^{ik}_{b}\mathcal{N}(\hat{g}^j_b|\mu^{ik}_{b},\Sigma^{ik}_{b})+\varepsilon),\\
\lambda^{ij}_G=\begin{cases}
    1, & \text{if $IoU > 0.5$}.\\
    0, & \text{otherwise}.
  \end{cases},~\hat{g}^{j}_{x}=\frac{(g^{j}_{x}-d^{i}_x)}{d^i_{w} },~\hat{g}^{j}_{y}=\frac{(g^{j}_{y}-d^{i}_y)}{d^i_{h}},
  \\
  \hat{g}^{j}_{w}=log(\frac{g^{j}_{w}}{d^{i}_w}),~\hat{g}^{j}_{h}=log(\frac{g^{j}_{h}}{d^{i}_h}),
\end{gathered}$}
\end{equation}

\noindent where $l$ is GMM parameters of bounding box ($\pi^{ik}_{b}$, $\mu^{ik}_{b}$, and $\Sigma^{ik}_{b}$), $N$ is the number of matched anchor boxes (called \textit{positive matches}), $K$ is the number of mixtures, $\lambda^{ij}_G$ is an indicator for matching the $i$-th anchor box $d^i_b$ to the $j$-th GT box of category $G$, and $\hat{g}^j_b$ is the $j$-th GT box. In experiments, we set $\varepsilon=10^{-9}$ for the numerical stability of the logarithm function.

\textbf{Classification:} We now focus on the classification head of the object detector. We model the output of every class as a GMM (see \fig{fig:network_re}a). Our approach estimates the mean $\hat{\mu}^k_{p}$ and variance $\hat{\Sigma}^k_{p}$ for each class, and the weights of the mixture $\hat{\pi}^k$ for each component of the GMM. We process the parameters of the GMM following Eq.~\ref{eq:GMMpreproc}, and obtain the class probability distribution for the $k$-th mixture by using the reparameterization trick~\cite{kingma2013auto} of applying Gaussian noise and variance $\Sigma^k_p$ to $\mu^k_p$~\cite{NIPS2017_7141}: 
\begin{equation}\small
\label{eq:reparameter}
\begin{aligned}
\hat{c}^k_{p}=\mu^k_{p}+\sqrt{\Sigma^k_{p}}\gamma,~~\gamma\sim \mathcal{N}(0, 1),
\end{aligned}
\end{equation}
\noindent where $\gamma$ is the auxiliary noise variable and has the same size as $\mu^k_{p}$ and $\Sigma^k_{p}$.

\textbf{Classification loss: }For training the mixture density network for classification, we propose a loss function that takes into account the IoU of the anchor box compared to the GT box and considers the hard negative mining. More precisely, we formulate the classification loss as a combination of two terms $L^{Pos}_{cl}$ and $L^{Neg}_{cl}$ representing the contribution of positive and negative matches:
\begin{equation}\small
\label{eq:cls_loss}
\begin{gathered}
L^{Pos}_{cl}(\lambda,c)=-\sum_{i\in{Pos}}^N\lambda^{ij}_G\sum_{k=1}^K\pi^{ik}(\hat{c}^{j}_G-log\sum_{p=0}^C e^{\hat{c}^{ik}_p})\\
L^{Neg}_{cl}(c)=-\sum_{i\in{Neg}}^{M\times{N}}\sum_{k=1}^K\pi^{ik}(\hat{c}^{i}_0-log\sum_{p=0}^Ce^{\hat{c}^{ik}_p}),
\end{gathered}
\end{equation}
\noindent where $N$ is the number of positive matches, $K$ is the number of mixtures, $C$ is the number of classes, with $0$ representing the background class $\hat{c}^{i}_0$, $\hat{c}^{j}_G$ is the GT class for the $j$-th GT box, $\hat{c}^{ik}_p$ is the result calculated by Eq.~\ref{eq:reparameter}, $\lambda^{ij}_G$ is the same as used in Eq.~\ref{eq:localization_loss}, and $M$ is the hard negative mining ratio. Instead of using all the negative matches, we sort them using the proposed mixture classification loss and choose top $M\times N$ as final negative matches for training. In experiments, we set $M$ to $3$ as suggested in~\cite{liu2016ssd}.

\textbf{Final loss: }We define the overall loss to train the object detector using mixture density network as:
\begin{equation}\small
\centering
\resizebox{.99\hsize}{!}{$
L=\begin{cases}
    \frac{1}{N}(L_{loc}(\lambda,l,g)/\eta+L^{Pos}_{cl}(\lambda,c)+L^{Neg}_{cl}(c)),&\text{if $N>0$}.\\
    0,&\text{otherwise}.
  \end{cases}$}
\end{equation}
\noindent where $N$ is the number of positive matches. In experiments, we set $\eta$ to $2$ as suggested in~\cite{choi2019gaussian}.

At inference, we can compute the coordinates of the bounding box $R_b$ and the confidence score for each class $P_i$ by summing the components of the mixture model as follows:
\begin{equation}\small
\begin{gathered}
\textrm{Localization:}~{R_b=\sum_{k=1}^K\pi^k_b\mu^k_b},\\
\textrm{Classification:}~{P_i=\sum_{k=1}^K\pi^k\frac{e^{\mu^k_i}}{\sum_{j=0}^Ce^{\mu^k_j}}}.
\end{gathered}
\end{equation}

\subsection{Improving parameter efficiency}
\label{sec-efficient}
%\textbf{Improving parameter efficiency:} 
In order to predict a probability distribution of the output values, our approach involves modifying the last layer of the network and therefore incurs an increase in the number of parameters, especially in the classification head. More precisely, for an output feature map of size $F\times F$, with $C$ classes, $D$ anchor boxes, and each bounding box defined using $4$ coordinates, the number of parameters in the new layer added to estimate a $K$-component GMM with $3$ parameters is $F\times F\times D\times (4 \times 3 \times K )$ for the localization and $F\times F\times D\times (C \times 2 \times K +K)$ for  classification. We see that the number of parameters in the classification head is proportional to the number of classes.

In this section, we focus on improving the efficiency of the algorithm by reducing the number of parameters in the classification head. To this end, as shown in~\fig{fig:network_re}b, we relax the problem of estimating the variance $\Sigma_p$, in order to reduce the number of parameters with $F\times F\times D\times (C \times K +K)$. Instead, we obtain class probabilities as $\hat{c}^k_{p}=Softmax(\mu^k_{p})$, and use them to estimate the aleatoric uncertainty as follows: 
\begin{equation}\small
%u_{al}=\sum_{i=1}^K\pi^{i}(diag(\hat{c}^{ik}_p)-(\mu^{i})^{\otimes2})\\
u_{al}=\sum_{k=1}^K\pi^{k}(diag(\hat{c}^{k}_{p})-(\hat{c}^{k}_{p})^{\otimes2}),%~~\hat{c}^j_{p}=\mu^j_{p}
\end{equation}
\noindent where $diag(q)$ is a diagonal matrix with the elements of the vector $q$ and $q^{\otimes2}=qq^T$. In this case, $u_{al}$ is $C\times C$ matrix where the value of each diagonal element can be interpreted as a class-specific aleatoric uncertainty~\cite{kwon2020uncertainty}. 
%where the diagonal elements represent the aleatoric uncertainty~\cite{kwon2020uncertainty}. 

Finally, we modify the classification loss for training the model with improved parameter efficiency as follows:
\begin{equation}\small
\label{eq:cls_loss_reduce}
\begin{gathered}
L^{Pos}_{cl}(\lambda,c)=-\sum_{i\in{Pos}}^N\lambda^{ij}_G\sum_{k=1}^K\pi^{ik}(\hat{c}^{j}_G-log\sum_{p=0}^C e^{\hat{\mu}^{ik}_p})\\
L^{Neg}_{cl}(c)=-\sum_{i\in{Neg}}^{M\times{N}}\sum_{k=1}^K\pi^{ik}(\hat{c}^{i}_0-log\sum_{p=0}^Ce^{\hat{\mu}^{ik}_p}),
\end{gathered}
\end{equation}
\noindent where all parameters are same as Eq.~\ref{eq:cls_loss}, except for class probability $\hat{\mu}^{ik}_p$.

\subsection{Scoring function}
The scoring function in active learning provides a single value per image indicating its informativeness. Our scoring function estimates the informativeness of an image by aggregating all the aleatoric and epistemic uncertainty values for each detected object in the image.  
%In our case, our model estimates, for each of the aleatoric and epistemic uncertainty, five uncertainty values per bounding box that will be aggregated over all the objects in the image to estimate its informativeness. 

Specifically, let $U=\{u^{ij}\}$ be the set of uncertainty values (aleatoric or epistemic) of a group of images where $u^{ij}$ is the uncertainty for the $j$-th object in the $i$-th image. For localization, $u^{ij}$ is the maximum value over the 4 bounding box outputs. We first normalize these values using z-score normalization ($\tilde{u}^{ij}=(u^{ij}-\mu_U)/\sigma_U$) to compensate for the fact that the values for the coordinates of the bounding box are unbounded and each uncertainty of an image might have a different range of values. We then assign to each image the maximum uncertainty over the detected objects $u^i= \max_j{\tilde{u}^{ij} }$. We empirically find that taking the maximum over the coordinates and the objects performs better than by taking the average.

Using the algorithm described above, we obtain four different normalized uncertainty values for each image: epistemic and aleatoric for classification and localization, $\mathbf{u}=\{u^i_{ep_{c}},u^i_{al_{c}},u^i_{ep_b},u^i_{al_b}\}$, respectively. The remaining part is to aggregate these scores into a single one. We explore different combinations of scoring functions that aggregate these uncertainties, including sum or taking the maximum, like other active learning studies~\cite{haussmann2020scalable,DBLP:conf/bmvc/RoyUN18}. As we will show in our experiments, taking the maximum over them achieves the highest results.
\section{Experiments}
In this section, we demonstrate the benefits of our approach. We first study the impact of using probabilistic modeling for the object detector and then analyze the proposed scoring function and relevant SOTA approaches in the context of active learning.  

\noindent \textbf{Datasets:} We use PASCAL VOC~\cite{pascal-voc-2012} and MS-COCO~\cite{lin2014microsoft} datasets. For PASCAL VOC, that contains $20$ object categories, we use VOC07 (VOC2007) \textit{trainval} and VOC07+12 \textit{trainval} (union of VOC2007 and VOC2012) for training and evaluate our results on VOC07 \textit{test}. For MS-COCO, that contains $80$ object categories, we use MS-COCO \textit{train2014} for training and evaluate our results on \textit{val2017}. 
\begin{table}[]
\centering
\resizebox{0.99\linewidth}{!}
{
\begin{tabular}{c|c|cc|cc}
\hline 
         &        & \multicolumn{2}{c|}{(a) \textbf{VOC07}}                   & \multicolumn{2}{c}{(b) \textbf{MS-COCO}}                 \\ \hline
Method   & Head   & IoU\textgreater0.5 & IoU\textgreater0.75 & IoU\textgreater0.5 & IoU\textgreater0.75 \\ \hline
SSD      & -      & 69.29$\pm$0.51       & 43.36$\pm$1.24        & 25.63$\pm$0.40       & 11.93$\pm$0.60        \\ \hline
SGM      & Loc    & 70.20$\pm$0.27       & 45.39$\pm$0.23        & 27.20$\pm$0.08       & 12.70$\pm$0.16        \\
MDN      & Loc    & 70.09$\pm$0.22       & 46.01$\pm$0.27        & 27.67$\pm$0.12       & 13.53$\pm$0.05        \\
SGM      & Cl     & 69.95$\pm$0.41       & 44.25$\pm$0.26        & 27.23$\pm$0.12       & 12.50$\pm$0.08        \\
MDN      & Cl     & 70.47$\pm$0.17       & 44.47$\pm$0.06        & 27.33$\pm$0.09       & 12.67$\pm$0.09        \\ \hline
$Ours_{gmm}$ & Loc+Cl & 70.19$\pm$0.36   & 46.11$\pm$0.38        & 27.70$\pm$0.08       & 13.57$\pm$0.19        \\
$Ours_{eff}$ & Loc+Cl & 70.45$\pm$0.06   & 46.18$\pm$0.26        & 27.33$\pm$0.04       & 13.33$\pm$0.12        \\ \hline \end{tabular}
}
\caption{mAP (in \%) of different instances of our approach compared to the original SSD network. $SGM$ and $MDN$ refer to single and multiple Gaussian models, and we apply those to localization (Loc), classification (Cl), and their combination (Loc+Cl).}
\label{various_instances}
\vspace{-0.25cm}
\end{table}

\noindent \textbf{Experimental settings:} 
%For fair comparison with other studies ~\cite{DBLP:conf/cvpr/YooK19, DBLP:conf/bmvc/RoyUN18}, we use Single Shot MultiBox Detector (SSD)~\cite{liu2016ssd}, with a VGG-16 backbone~\cite{vgg}. We follow all the implementation details of \cite{DBLP:conf/cvpr/YooK19}. 
We employ Single Shot MultiBox Detector (SSD)~\cite{liu2016ssd}, which is widely used in active learning studies~\cite{DBLP:conf/bmvc/RoyUN18,DBLP:conf/cvpr/YooK19}, with a VGG-16 backbone~\cite{vgg}. We train our models for $120,000$ iterations using SGD with a batch size of $32$ and a maximum learning rate of $0.001$. We use a learning rate warm-up strategy for the first $1,000$ iterations and divide the learning rate by $10$ after $80,000$ and $100,000$ iterations. We set the number of Gaussian mixtures to $4$, and in the supplementary materials, we provide an ablation study on the number of mixtures. Unless specified otherwise, we report the performance using the average and standard deviation of mAP for three independent trials.

\subsection{Object detection with probabilistic modeling}
\label{4.1}
We first analyze the impact of using our proposed probabilistic modeling for object detection on PASCAL VOC and MS-COCO. For MS-COCO, we use a random subset of $5,000$ training images from \textit{train2014}. We compare the accuracy of our GMM $Ours_{gmm}$ and the model with improved parameter efficiency $Ours_{eff}$ to the SSD~\cite{liu2016ssd} and several network configurations either using single or multiple Gaussians for the classification or localization heads. %For the evaluation, we provide the average mAP of three experiments with the standard metric (IoU$>$0.5) and the strict metric (IoU$>$0.75).
%$Ours$ and $Ours^*$ in the Table~\ref{various_instances}a and~\ref{various_instances}b are the proposed GMM and the model that improves the parameter efficiency,
\begin{figure}[!t]
	\centering
	%\hspace{-0.0cm}
	\begin{tabular}{cc}
	\hspace{-0.0cm}\includegraphics[scale=0.29]{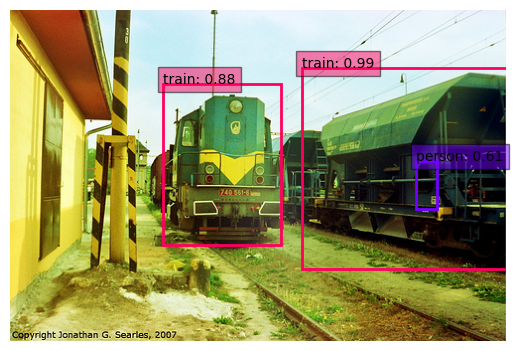}&
	\hspace{-0.2cm}\includegraphics[scale=0.29]{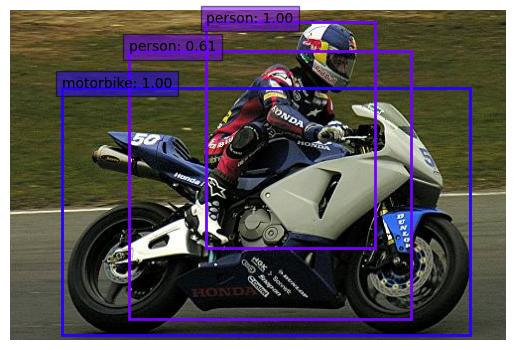}\\
	\hspace{-0.0cm}\scriptsize{\begin{tabular}{ll}$u_{al_b}$: \textbf{3.60} & $u_{al_c}$: $0.96$\\ $u_{ep_b}$: $1.06$ & $u_{ep_c}$: $-0.19$ \end{tabular}}&
	\hspace{-0.2cm}\scriptsize{\begin{tabular}{ll}$u_{al_b}$: $1.71$ & $u_{al_c}$: $-0.50$ \\ $u_{ep_b}$: \textbf{11.45}& $u_{ep_c}$: $-0.38$ \end{tabular}}\\

	\hspace{-0.0cm}\includegraphics[scale=0.29]{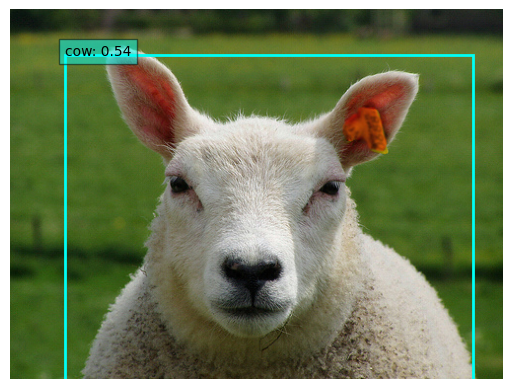}&
	\hspace{-0.2cm}\includegraphics[scale=0.29]{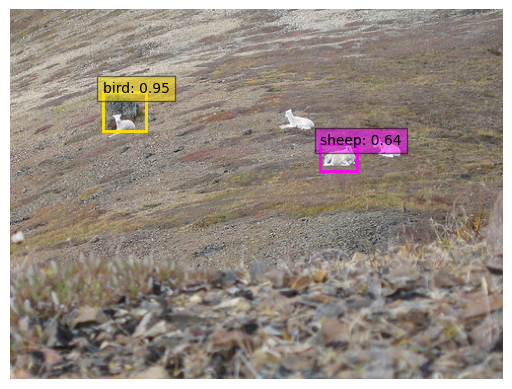}\\
	\hspace{-0.0cm}\scriptsize{\begin{tabular}{ll}$u_{al_b}$: $-1.09$ & $u_{al_c}$: \textbf{8.80}\\ $u_{ep_b}$: $-0.38$ & $u_{ep_c}$: $1.35$  \end{tabular}}&
	\hspace{-0.6cm}\scriptsize{\begin{tabular}{ll}$u_{al_b}$: $0.74$ & $u_{al_c}$: $1.06$\\ $u_{ep_b}$: $0.80$ & $u_{ep_c}$: \textbf{7.14}\end{tabular}}%\\	(a)&(b)&(c)&(d)
	\end{tabular}
	\vspace{-.2cm}
	\caption{Examples of aleatoric and epistemic uncertainties for inaccurate detections, see more examples in the supplementary material. Starting from top-left image and going in clockwise direction: Person is a false positive; Person bounding box is not correct; A sheep is misclassified as a bird; A sheep is misclassified as a cow.}
	\label{fig:convent_out}
	\vspace{-.7cm}
\end{figure}
%Our loss function serves as a regularizer for inconsistent data, thus allowing the model to focus on consistent data~\citep{choi2019gaussian}

In \tab{various_instances}a and \tab{various_instances}b, we summarize the results of this experiment performed on VOC07 and MS-COCO, respectively. As shown, all networks that include probabilistic modeling outperform the SSD on both datasets. 
%We also observe that the accuracy is higher for models using a mixture density network compared to using a single Gaussian. This is because of the regularization effect of our proposed loss function, which has a loss attenuation due to aleatoric uncertainty~\cite{choi2019gaussian}.
This is because of the regularization effect of the proposed loss function, which has a loss attenuation due to aleatoric uncertainty~\cite{choi2019gaussian}. As a result, we obtain models that are robust to noisy data. Considering both the normal (IoU$>$0.5) and the strict metric (IoU$>$0.75), $Ours_{gmm}$ and $Ours_{eff}$ outperform all other variations on VOC07. On MS-COCO, $Ours_{gmm}$ outperforms all other instances and the baseline, while $Ours_{eff}$ reaches competitive results. We expect that the amount of noisy data in MS-COCO is larger than that of PASCAL VOC because MS-COCO has more diverse data. As shown in Eq.~\ref{eq:cls_loss_reduce}, there is no aleatoric uncertainty in $Ours_{eff}$'s classification loss and therefore, we argue that the regularization by aleatoric uncertainty has a greater effect in MS-COCO.

In \fig{fig:convent_out}, we present representative examples of uncertainty scores for several images where the detector fails to detect the object. As shown, each uncertainty value (bold numbers in~\fig{fig:convent_out}) provides a different insight into some particular failure. Localization uncertainties are related to the accuracy of the bounding box prediction, whereas classification uncertainties are related to the accuracy of the category prediction. Interestingly, in these examples, even if the predictions are wrong, uncertainty values seem to be uncorrelated suggesting each uncertainty could predict inaccurate results independently. From these results, we can conclude that the proposed approach not only computes uncertainty in a single forward pass of a single model but also boosts the performance of the detection network. As shown in the next experiment, combining these values will improve the data selection process during active learning.

% \begin{figure*}[h!]
% 	\centering
% 	\begin{tabular}{ccc}
% 	\includegraphics[width=0.33\linewidth]{fig/single_new_1k.pdf} \vspace{-.03cm}&
% 	\includegraphics[width=0.33\linewidth]{fig/multi_new_1k.pdf} \vspace{-.03cm}&
% 	\includegraphics[width=0.26\linewidth]{fig/compute_new.pdf}
% 	\vspace{-.03cm}
% 	\\	\vspace{-.13cm}
% 	\small{(a)}&\small{(b)}&\small{(c)}
% 	\end{tabular}
% 	%\vspace{-.2cm}
% 	\caption{\textbf{VOC07+12:} %Active learning results of object detection.
% 	a) Comparison to published works using a single model for scoring. Numbers taken from~\cite{DBLP:conf/cvpr/YooK19}; b) Comparison with multiple model-based methods, ensemble and MC-dropout. Numbers we reproduced; c) Model parameters in millions (M) and forward time in seconds (sec). All numbers used to create the plot are summarized in the supplementary material.}
% 	\label{fig:result_0712}
% 	%\vspace{-.5cm}
% \end{figure*}

\subsection{Active learning evaluation}
We now focus on evaluating the performance of our active learning on PASCAL VOC and MS-COCO datasets. We use an initial set of $2,000$ for VOC07, $1,000$ for VOC07+12 as suggested by ~\cite{DBLP:conf/cvpr/YooK19}, and $5,000$ samples in MS-COCO as suggested by ~\cite{DBLP:conf/accv/KaoLS018}. Then, during the active learning stage, for each unlabeled image, we apply non-maximum suppression and we compute the uncertainties for each of the \textit{surviving} objects. The scoring function aggregates these uncertainties using the maximum or sum to provide the final informativeness score for the image. 
We score the set of unlabeled images and select the $1,000$ images~\cite{DBLP:conf/cvpr/YooK19} with the highest score. Then, we add them to the labeled training set and repeat this process for several active learning cycles. For every active learning iteration, we train the model from scratch, using ImageNet pretrained weight. 

\noindent \textbf{Scoring aggregation function:}
\begin{table}[t]
\centering
% \vspace{-1.45cm}
\centering
\resizebox{0.99\linewidth}{!}
{
\begin{tabular}{c|ccc}
\hline
 \multirow{2}{*}{\begin{tabular}[c]{@{}c@{}}Aggregation\\function\end{tabular}}  &\multicolumn{3}{c}{mAP in \% (\# images)}\\
      & 1st (2k)  &  2nd (3k)   &  3rd (4k)         \\ \hline
$random$ $sampling$    & 62.43$\pm$0.10      & 66.36$\pm$0.13    & 68.47$\pm$0.09           \\\hline
$u_{al_b}$ & 62.43$\pm$0.10     & 67.06$\pm$0.18     & 68.84$\pm$0.18            \\ %\hline
$u_{ep_b}$ & 62.43$\pm$0.10      & 66.75$\pm$0.26     & 69.01$\pm$0.17           \\ %\hline
$u_{al_c}$ & 62.43$\pm$0.10      & 67.09$\pm$0.09     & 68.75$\pm$0.08           \\ %\hline
$u_{ep_c}$ & 62.43$\pm$0.10      & 66.51$\pm$0.12     & 68.95$\pm$0.13           \\ \hline
$\sum_{j\in\{{al_b},{ep_b}\}} u_j$  & 62.43$\pm$0.10   & 67.01$\pm$0.10 & 68.58$\pm$0.29    \\ %\hline
$\sum_{j\in\{{al_c},{ep_c}\}} u_j$  & 62.43$\pm$0.10   & 67.07$\pm$0.27 & 69.03$\pm$0.20    \\ \hline
$\sum_{j\in\{{al_b},{al_c}\}} u_j$  & 62.43$\pm$0.10   & 66.96$\pm$0.08 & 68.92$\pm$0.23    \\ %\hline
$\sum_{j\in\{{ep_b},{ep_c}\}} u_j$  & 62.43$\pm$0.10   & 66.49$\pm$0.14 & 68.62$\pm$0.24    \\ \hline
$\sum_{j\in\{{al_b},{ep_b},{al_c},{ep_c}\}} u_j$ & 62.43$\pm$0.10   & 67.04$\pm$0.28 & 69.09$\pm$0.30 \\ \hline
$\max_{j\in\{{al_b},{ep_b}\}} u_j$  & 62.43$\pm$0.10   & 66.82$\pm$0.21   & 68.95$\pm$0.22    \\ %\hline
$\max_{j\in\{{al_c},{ep_c}\}} u_j$  & 62.43$\pm$0.10   & 66.87$\pm$0.14   & 68.99$\pm$0.31    \\ \hline
$\max_{j\in\{{al_b},{al_c}\}} u_j$  & 62.43$\pm$0.10   & 67.18$\pm$0.10 & 69.06$\pm$0.25 \\ %\hline
$\max_{j\in\{{ep_b},{ep_c}\}} u_j$  & 62.43$\pm$0.10   & 66.72$\pm$0.10 & 68.99$\pm$0.21     \\ \hline
$\max_{j\in\{{al_b},{ep_b},{al_c},{ep_c}\}} u_j$ & 62.43$\pm$0.10   & \textbf{67.32}$\pm$0.12 & \textbf{69.43}$\pm$0.11       \\ \hline
\end{tabular}
}
\caption{\textbf{VOC07:} Comparison of scoring aggregation functions for active learning based on the aleatoric uncertainty, epistemic uncertainty, and their combination of each task.}
\label{uncertainty_ablation}
\vspace{-.3cm}
\end{table}
We compare the active learning performance obtained using different functions to aggregate the aleatoric and epistemic uncertainties of both classification and localization heads. In particular, we compare seven different instances of our approach with random sampling: 1) Only the aleatoric or epistemic uncertainty on each task; 2) The sum of aleatoric and epistemic uncertainty on the localization or classification head; 3) The sum of aleatoric or epistemic uncertainty on the localization and classification; 4) The sum of aleatoric and epistemic uncertainties for both localization and classification; 5) The maximum of aleatoric and epistemic uncertainty on the localization or classification head; 6) The maximum of aleatoric or epistemic uncertainty on the localization and classification, and 7) The maximum value of these four uncertainties. The results for this comparison are shown in \tab{uncertainty_ablation}. Our approach using the maximum value of aleatoric and epistemic uncertainties of both localization and classification tasks consistently outperforms all the other aggregation functions on each active learning iterations. More concretely, the maximum value of all uncertainties for both tasks shows better data selection performance in active learning than others. Based on these results, we use the maximum value of all uncertainties as a scoring function during active learning to compare with other active learning studies.

In \tab{overlap_ratio}, we summarize the overlap in the selection as a function of the uncertainty measure. The overlapping ratio using both uncertainties is $48\%$ for localization and $33\%$ for classification. More importantly, when we consider both uncertainties on localization and classification together, the overlapping ratio decreases to barely $14\%$. This suggests that uncertainty measures obtained for localization and classification are diversified and their combination improves the image selection process.

\begin{table}[t]
\centering
\resizebox{0.78\linewidth}{!}{
\begin{tabular}{c|cc|cc}
\hline
 & \multicolumn{2}{c|}{Localization} & \multicolumn{2}{c}{Classification} \\
 & Aleatoric     & Epistemic  & Aleatoric    & Epistemic  \\
 &  $u_{al_b}$     & $u_{ep_b}$    &  $u_{al_c}$   &   $u_{ep_c}$  \\ \hline
$u_{al_b}$ & 100   & 48    & 6      & 11            \\
 $u_{ep_b}$ & 48   & 100   & 7      & 14            \\ \hline
$u_{al_c}$ & 6     & 7     & 100    & 33            \\
$u_{ep_c}$ & 11    & 14    & 33     & 100           \\ \hline
\end{tabular}
}
\caption{Overlapping ratio (in \%) of selected images as a function of the type of uncertainty used.}
\label{overlap_ratio}
%\vspace{-.15cm}
\end{table}
\begin{table}
\centering
\resizebox{0.99\linewidth}{!}{
\centering
\begin{tabular}{c|ccc|c|c}
\hline 
                        & \multicolumn{3}{c|}{mAP in \% (\# images)} & Number of      & Forward    \\
                        & 1st (2k)     & 2nd (3k)     & 3rd (4k)     & para. ($\times10^6$) & time (sec) \\ \hline
Random~\cite{liu2016ssd}                     & 62.43$\pm$0.10      & 66.36$\pm$0.13      & 68.47$\pm$0.09     & 52.35           & 0.031      \\
Entropy~\cite{DBLP:conf/bmvc/RoyUN18}        & 62.43$\pm$0.10      & 66.85$\pm$0.12      & 68.70$\pm$0.18     & 52.35           & 0.031      \\
Core-set~\cite{DBLP:conf/iclr/SenerS18}      & 62.43$\pm$0.10      & 66.57$\pm$0.20      & 68.57$\pm$0.26     & 52.35           & 0.031      \\
LLAL~\cite{DBLP:conf/cvpr/YooK19}            & 62.47$\pm$0.16      & 67.02$\pm$0.11      & 68.90$\pm$0.15     & 52.71           & 0.036      \\ \hline
MC-dropout~\cite{feng2019deep}               & 62.43$\pm$0.19      & 67.10$\pm$0.07      & 69.39$\pm$0.09     & 52.35           & 0.689          \\
Ensemble~\cite{haussmann2020scalable}        & 62.43$\pm$0.10      & 67.11$\pm$0.26      & 69.26$\pm$0.14     & 157.05          & 0.093      \\ \hline
$Ours_{gmm}$   & 62.43$\pm$0.10      & 67.32$\pm$0.12      & 69.43$\pm$0.11     & 52.35           & 0.031      \\
$Ours_{eff}$   & 62.91$\pm$0.16      & 67.61$\pm$0.17      & 69.66$\pm$0.17     & 41.12           & 0.029      \\\hline 
\end{tabular}
}
\caption{\textbf{VOC07:} Comparison of mAP and computing cost of active learning with most relevant approaches. \textit{Para.} and \textit{sec} refer to \textit{parameters} and \textit{seconds}, respectively.}
\label{al_07}
\vspace{-.25cm}
\end{table}

\begin{figure*}[!t]
	\centering
	\begin{tabular}{ccc}
	\includegraphics[width=0.39\linewidth]{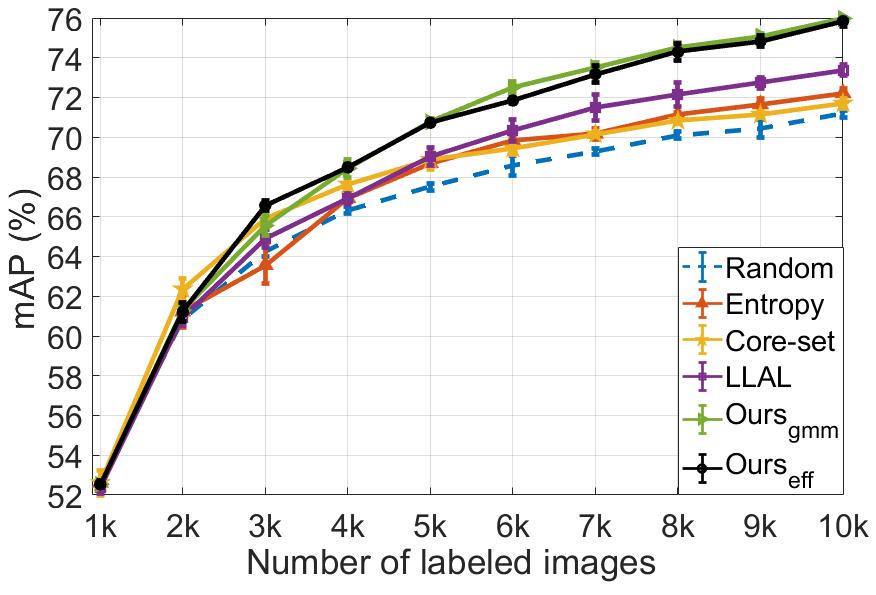} \vspace{-.05cm}&
	\includegraphics[width=0.39\linewidth]{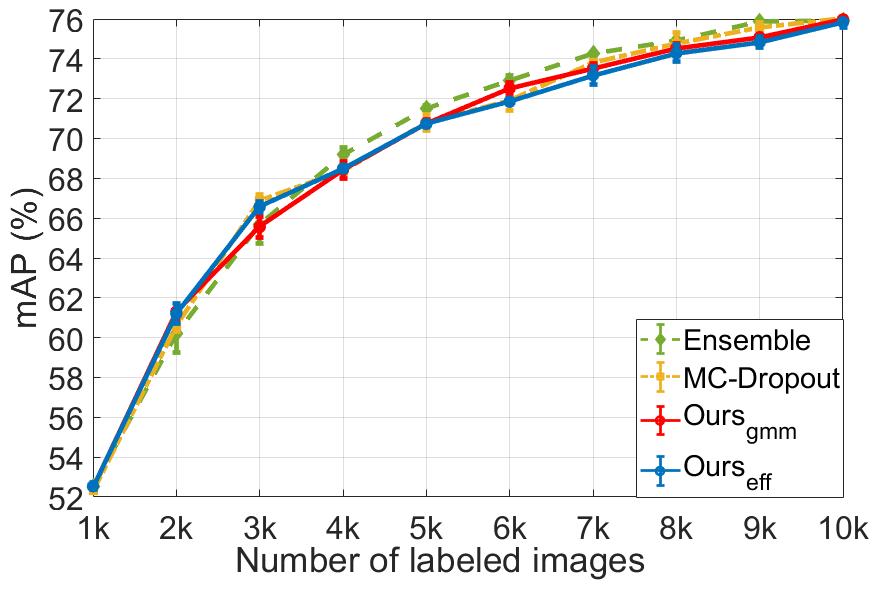} \vspace{-.05cm}&
	\\	\vspace{-.2cm}
	\small{(a)}&\small{(b)}
	\end{tabular}
	\vspace{-.15cm}
	\caption{\textbf{VOC07+12:} %Active learning results of object detection.
	a) Comparison to published works using a single model for scoring. Numbers are taken from~\cite{DBLP:conf/cvpr/YooK19}; b) Comparison to multiple model-based methods, ensemble and MC-dropout. Details of the numbers to reproduce the plot are in the supplementary material.}
	\label{fig:result_0712}
	\vspace{-0.3cm}
\end{figure*}

\begin{figure}[!t]
	\centering
	\includegraphics[width=0.72\linewidth]{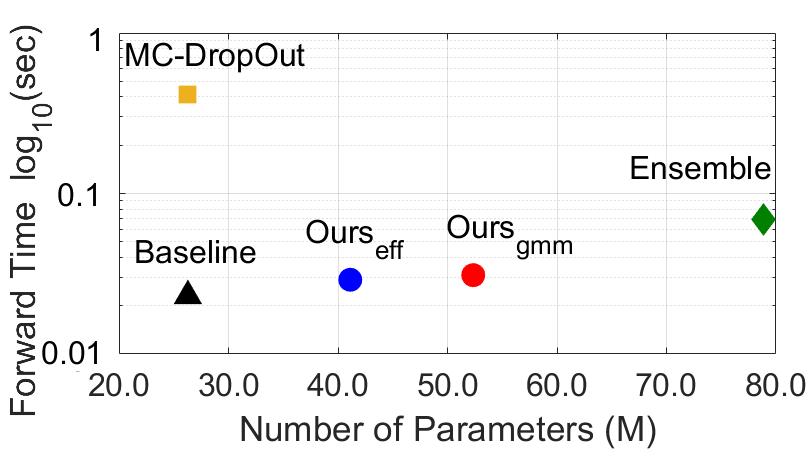}
	\vspace{-.1cm}
	\caption{\textbf{VOC07+12:} Computing cost comparison to baseline and methods using multiple models; Model parameters in millions (M) and the forward time in seconds (sec). See the supplementary material for numerical details.}
	\label{fig:result_0712_compute}
	\vspace{-.4cm}
\end{figure}
\noindent \textbf{Comparison to SOTA on VOC07:} In \tab{al_07}, we summarize the active learning results and computing cost of our method compared to relevant active learning approaches in the literature. In order to compare the computing cost, we provide the number of parameters and the forward time of each method. In general, a fast forward (backward) step and a small model size leads to a lower training cost and data sampling time during active learning ~\cite{highlander2016very,iandola2016squeezenet}. 

To focus on the active learning, we reproduce all numbers by applying each sampling method to the proposed GMM architecture where the output is a mixture distribution (\textit{i.e.}, the same model as $Ours_{gmm}$). For ensembles, we follow~\cite{DBLP:conf/cvpr/BeluchGNK18}, building an ensemble of three independent models. For MC-dropout, we add dropout layers with $p=0.1$ to the six convolutional layers composing the extra feature layers in SSD. We compute the image scores using $25$ forward passes~\cite{DBLP:conf/cvpr/BeluchGNK18}. For these two methods and entropy-based method, we follow the most common approach in the literature and estimate the final image score as the average entropy on the classification head~\cite{haussmann2020scalable}. For core-set~\cite{DBLP:conf/iclr/SenerS18}, we follow~\cite{DBLP:conf/cvpr/YooK19}, using features of fully connected layer-7 in VGG-16. For LLAL~\cite{DBLP:conf/cvpr/YooK19}, we implement the \textit{learning loss prediction module} on the proposed GMM architecture. As a baseline, we use random sampling on our GMM architecture. Note, we train all methods with exactly the same hyperparameters as mentioned in the \textit{experimental settings}. 
%As shown in \tab{al_07}, both instances of our approach consistently outperform all the other methods in every active learning iteration. 
As shown in \tab{al_07}, both instances of our approach consistently outperform all the other single-model based methods~\cite{liu2016ssd, DBLP:conf/bmvc/RoyUN18, DBLP:conf/iclr/SenerS18, DBLP:conf/cvpr/YooK19} in every active learning iteration. Compared to multi-model based methods~\cite{feng2019deep,haussmann2020scalable}, our proposed methods show higher accuracy while requiring a significantly lower computing cost. These results demonstrate that despite having a lower computational cost, our proposed method improves the active learning sampling performance compared to previous works.

\noindent \textbf{Comparison to SOTA on VOC07+12:}
We now compare our approach to existing single-model based approaches on VOC07+12. Here, we consider the SOTA results reported in~\cite{DBLP:conf/cvpr/YooK19} including LLAL~\cite{DBLP:conf/cvpr/YooK19} and core-set~\cite{DBLP:conf/iclr/SenerS18}, in addition to entropy~\cite{DBLP:conf/bmvc/RoyUN18} and random sampling. We use the same open-source and setting used in~\cite{DBLP:conf/cvpr/YooK19} for a fair comparison. To solely focus on active learning, we compare the performance based on the same baseline with~\cite{DBLP:conf/cvpr/YooK19}, \textit{i.e.}, SSD. To do this, we train the SSD using a dataset sampled by our proposed scoring function in the proposed GMM architecture and the architecture with improved parameter efficiency. To verify the influence of the initial training set for comparison, we run $5$ independent trials with different seeds for the initial choice of the labeled set. We then obtain an average mAP of $0.5246$ with a standard deviation of $0.003$ that suggests little variations when experiments use a different initial subset of images. As shown in \fig{fig:result_0712}a, our method outperforms all the other single-model based methods. In the last active learning iteration, our approach achieves $0.7598$ mAP which is $2.6$ percent points higher than the score achieved by LLAL~\cite{DBLP:conf/cvpr/YooK19} ($0.7338$ mAP), thus showing a high-performance improvement in active learning based on a single model.

%we follow the same experimental setting mentioned in~\tab{al_07} and apply it to the SSD.

Finally, we compare our approach with methods using multiple models, \ie, ensembles~\cite{haussmann2020scalable} and MC-dropout~\cite{feng2019deep}. For ensembles and MC-dropout, we follow the same design mentioned in~\tab{al_07} and apply it to the SSD. In \fig{fig:result_0712}b and \fig{fig:result_0712_compute}, we present the accuracy and computational cost comparison of these methods. As shown in \fig{fig:result_0712}b, in terms of the accuracy, our approach performs on par with MC-dropout and ensembles. However, our method uses a single forward pass of a single model to estimate the uncertainties, which is more efficient than ensembles and MC-dropout based methods. With respect to the number of parameters, MC-dropout has the same number of parameters as SSD since dropout layers do not add any new parameters, but it requires multiple forward passes. Our approach adds extra parameters for the estimation of two types of uncertainty to the last layer of each head and therefore, the number of parameters is larger than in SSD. In ensemble-based methods, the number of parameters is proportional to the number of SSD models in the ensemble~\cite{lakshminarayanan2017simple}. As shown in~\fig{fig:result_0712_compute}, our method requires significantly less computing cost than MC-dropout and ensemble-based method. In summary, our method provides the best trade-off between accuracy and computing cost for active learning.

\begin{table}
\centering
\resizebox{0.99\linewidth}{!}{
\centering
\begin{tabular}{c|ccc|c|c}
\hline 
                        & \multicolumn{3}{c|}{mAP in \% (\# images)} & Number of      & Forward    \\ 
                        & 1st (5k)     & 2nd (6k)     & 3rd (7k)     & para. ($\times10^6$) & time (sec) \\ \hline
Random~\cite{liu2016ssd}                  & 27.70$\pm$0.08    & 28.70$\pm$0.13     & 29.83$\pm$0.04     & 116.51        & 0.152      \\
Entropy~\cite{DBLP:conf/bmvc/RoyUN18}     & 27.70$\pm$0.08    & 28.93$\pm$0.11     & 29.89$\pm$0.09     & 116.51        & 0.152      \\
Core-set~\cite{DBLP:conf/iclr/SenerS18}   & 27.70$\pm$0.08    & 28.99$\pm$0.01     & 29.93$\pm$0.06     & 116.51        & 0.152      \\
LLAL~\cite{DBLP:conf/cvpr/YooK19}         & 27.71$\pm$0.03    & 28.71$\pm$0.06     & 29.53$\pm$0.15     & 116.87        & 0.194      \\ \hline
MC-dropout~\cite{feng2019deep}            & 27.70$\pm$0.10    & 29.20$\pm$0.09     & 30.30$\pm$0.08     & 116.51        & 3.718      \\
Ensemble~\cite{haussmann2020scalable}     & 27.70$\pm$0.08    & 29.03$\pm$0.07     & 30.02$\pm$0.06     & 349.53        & 0.456      \\ \hline
$Ours_{gmm}$    & 27.70$\pm$0.08    & 29.28$\pm$0.05     & 30.51$\pm$0.12     & 116.51        & 0.152      \\
$Ours_{eff}$    & 27.33$\pm$0.04    & 29.06$\pm$0.08     & 30.02$\pm$0.05     & 73.20         & 0.141      \\ \hline 
\end{tabular}
}
\vspace{-0.2cm}
\caption{\textbf{MS-COCO:} Comparison of mAP and computing cost of active learning with most relevant methods. \textit{Para.} and \textit{sec} refer to \textit{parameters} and \textit{seconds}, respectively.}
\label{fig:al_coco_result}
%\vspace{-0.1cm}
\end{table}
\noindent \textbf{Comparison to  SOTA on MS-COCO:} In \tab{fig:al_coco_result}, we summarize the active learning performance and computing cost of our approaches compared to active learning methods in the literature.  To solely focus on the active learning, we reproduce all numbers by applying each sampling method to the proposed GMM architecture (\textit{i.e.}, the same model as $Ours_{gmm}$). For all methods, we follow the same settings as on~\tab{al_07}. As shown, both instances of our approach consistently outperform all the other single-model based methods~\cite{liu2016ssd, DBLP:conf/bmvc/RoyUN18, DBLP:conf/iclr/SenerS18, DBLP:conf/cvpr/YooK19} in each active learning cycle. In particular, LLAL~\cite{DBLP:conf/cvpr/YooK19} shows similar accuracy to random sampling on MS-COCO, because it does not take into account the large diversity of the data and the large number of classes present in the dataset. However, our approach also shows high accuracy on MS-COCO. Compared to multiple model-based methods~\cite{feng2019deep,haussmann2020scalable}, both instances of our approach require a much less computing cost while $Ours_{gmm}$ outperforms those methods, and $Ours_{eff}$ shows competitive results with a much lower computational cost. These results suggest that our approach generalizes to larger dataset that have a larger number of classes.

\begin{table}[t]
\centering
\resizebox{0.88\linewidth}{!}
{
\begin{tabular}{cc|c|cc}
\hline
 &                       & Baseline~\cite{ren2015faster} & $Ours_{gmm}$ & $Ours_{eff}$  \\ \hline
\multirow{2}{*}{\begin{tabular}[c]{@{}c@{}}mAP\\ (\%)\end{tabular}} 
& IoU\textgreater0.5  & 75.31$\pm$0.22    & 75.90$\pm$0.09  & 75.80$\pm$0.15  \\
& IoU\textgreater0.75 & 48.70$\pm$0.11    & 49.36$\pm$0.07  & 49.83$\pm$0.30  \\ \hline
\multicolumn{2}{c|}{\# of parameters (M)}
& 41.17    & 42.23 & 41.61      \\ \hline
\multicolumn{2}{c|}{Forward time (sec)}
& 0.059    & 0.062 & 0.060      \\ \hline
\end{tabular}
}
\vspace{-0.2cm}
\caption{\textbf{VOC07:} Performance comparison of our mixture models based on Faster-RCNN and the original Faster-RCNN as a baseline~\cite{ren2015faster}.}
\label{fig:scale}
\vspace{-0.2cm}
\end{table}
\begin{table}[t]
\centering
\resizebox{0.99\linewidth}{!}
{
\begin{tabular}{c|c|c|cc}
\hline 
&    & \multicolumn{3}{c}{mAP in \%} \\ %\hline 
Model & Backbone  & 
\begin{tabular}[c]{@{}c@{}}Random\\ selection\end{tabular} & 
\begin{tabular}[c]{@{}c@{}}$Ours_{gmm}$\\ selection\end{tabular} & 
\begin{tabular}[c]{@{}c@{}}$Ours_{eff}$\\ selection\end{tabular} \\ \hline
\multirow{3}{*}{SSD~\cite{liu2016ssd}} & VGG-16~\cite{vgg}     & 67.77$\pm$0.12      & 68.71$\pm$0.18     & 68.48$\pm$0.31  \\
                     & Resnet-34~\cite{DBLP:conf/cvpr/HeZRS16} & 65.53$\pm$0.17      & 67.00$\pm$0.14     & 67.20$\pm$0.13  \\
                     & Resnet-50~\cite{DBLP:conf/cvpr/HeZRS16} & 64.28$\pm$0.39      & 65.73$\pm$0.32     & 65.81$\pm$0.21  \\ \hline
Faster-RCNN~\cite{ren2015faster}  & Resnet-50-FPN~\cite{lin2017feature}  & 72.93$\pm$0.41   & 73.60$\pm$0.18  & 75.45$\pm$0.30          \\ \hline 
\end{tabular}
}
\vspace{-0.1cm}
\caption{\textbf{VOC07:} Transferability of a dataset created using the proposed scoring function and the mixture-based density models. As shown, datasets acquired using our method not only boost the performance of models using a different backbone but also the performance of  two-stage detector such as Faster-RCNN. }
\label{fig:transfer}
\vspace{-0.3cm}
\end{table}

\subsection{Scalability and dataset transferability}
Our method is not limited to single-stage detectors. Here, in a first experiment, we show how our method can be applied to a two-stage detector such as Faster-RCNN~\cite{ren2015faster} with FPN~\cite{lin2017feature}. For this experiment, we use the same PASCAL VOC dataset as in~\tab{various_instances}a. In \tab{fig:scale}, we show the summary of the accuracy and the computing cost of our mixture models based on Faster-RCNN and the original Faster-RCNN as a baseline. As shown, both versions of our approach outperform the original model with up to $1.13$ mAP improvement. Importantly, in this case, our approach is applied to the output layer of the detection network after region proposal in Faster-RCNN, therefore there is a negligible increase in computing cost and latency because the computation does not include the number of anchor boxes.

Finally, we study the transferability of actively acquired datasets. We compare the performance of SSD using different backbones such as Resnet-34 and Resnet-50~\cite{DBLP:conf/cvpr/HeZRS16}, and Faster-RCNN~\cite{ren2015faster} detector trained using our actively sampled dataset. We perform the experiments in the actively sampled dataset from the last active learning cycle in~\tab{al_07}. For completeness, we also report the accuracy obtained using random sampling. We summarize the results of this experiment in \tab{fig:transfer}. As shown, networks trained using the actively sampled dataset outperform those trained using random sampling with up to $2.52$ mAP improvement. Conclusively, our method not only scales to other object detection networks but also datasets actively acquired using our approach can be used to train other architectures.

\section{Conclusions}
We have proposed a novel deep active-learning approach for object detection. Our approach relies on mixture density networks to estimate, in a single forward pass of a single model, two types of uncertainty for both localization and classification tasks, and leverages them in the scoring function. Our proposed probabilistic modeling and scoring function achieve outstanding performance gains in accuracy and computing cost. We present a wide range of experiments on two publicly available datasets, PASCAL VOC and MS-COCO. Besides, our results suggest that our approach scales to new models with different architectures.

% We have proposed a novel deep active-learning approach for object detection. Our approach relies on mixture density networks to provide, in a single forward pass of a single model, probabilistic distribution for every output of the model. Our method efficiently estimates the aleatoric and epistemic uncertainty for each of these outputs. We aggregate all uncertainties from both the classification and localization heads of the model in active learning, which is crucial to improving performance. We have demonstrated the efficacy of our approach in two publicly available datasets, PASCAL VOC and MS-COCO. Our mixture model-based object detection network yields up to $2.8\%$ mAP improvements compared to baselines. Moreover, for active learning, our approach outperforms state-of-the-art active learning methods using a single model by up to 2.6\% mAP improvement and, importantly, performs on par with multiple model-based methods, but requires a significant reduction in computational cost. Besides, our results suggest that datasets obtained using active learning can be effectively used for training new models with different architectures.

\appendix

\section{Parameter Sensitivity}
\subsection{Accuracy as a function of $K$} 
In the main paper, we presented experiments using $K$=4 as the number of components in the mixture model. In~\tab{varying_k} we analyze the sensitivity of our results with respect to the number of components in the GMM. Specifically, we provide numbers for $K$=1, 2, 4, and 8. As in the main paper, we repeat the experiment three times and provide the average mAP and standard deviation for the normal (IoU$>$0.5) and the strict metric (IoU$>$0.75). We also provide the number of parameters and the forward time for each of these instantiations. As shown, the accuracy of $K$=1 is much lower than that of $K$=4, especially for IoU$>$0.75. Moreover, for $K$=1, epistemic uncertainty cannot be estimated (see Eq.1 in the main paper). The accuracy remains stable for other configurations with minor variations in mAP. However, as the number of parameters is proportional to $K$, there are significant variations in terms of the number of parameters and forward time. Given these results, we selected $K$=4 as a good trade-off between accuracy (normal and strict metric) and computing cost. In practice, the larger $K$, the more difficult to train the GMM due to fluctuation. This would be the reason for a drop in accuracy when $K$=8.

\begin{table}[h]
\centering
\resizebox{0.99\linewidth}{!}
{
\begin{tabular}{c|cc|c|c}
\hline
 $K$ & \multicolumn{2}{c|}{mAP (\%)}     & \multirow{2}{*}{\# of parameters (M)} & \multirow{2}{*}{Forward time (sec)} \\
\multirow{1}{*}{\# of mixture}                               & IoU\textgreater0.5 & IoU\textgreater0.75 &                                       &                                   \\ \hline
1      & 69.89$\pm$0.23       & 45.18$\pm$0.24         & 29.6           & 0.021 \\
2      & 70.29$\pm$0.29       & 45.98$\pm$0.38         & 37.6           & 0.025 \\
4      & 70.19$\pm$0.36       & 46.11$\pm$0.38         & 52.3           & 0.031 \\
8      & 70.01$\pm$0.29       & 45.69$\pm$0.28         & 81.8           & 0.051 \\ \hline
\end{tabular}
}
\caption{\textbf{VOC07:} mAP and computing cost as a function of the number of components in the mixture model. Model parameters in millions (M) and forward time in seconds (sec).}
\label{varying_k}
\vspace{-6pt}
\end{table}

\subsection{Accuracy as a function of input image resolution}
In order to check for the robustness of our method with respect to the image size, here we compare the performance of the network trained using higher resolution images (512$\times$512). The experiment is analogous to the experiment we showed in Tab.1a in the main paper. We compare the results of SSD~\cite{liu2016ssd}, with the results of our method. As we can see in~\tab{SSD512}, as expected, increasing the resolution of the input image yields a significant improvement in mAP score for all the methods. For high-resolution input images, our method outperforms SSD in the normal metric (IoU$>$0.5) by $0.51$ percent points (\textit{pp}), and shows significant improvement when evaluated in the strict metric (IoU$>$0.75), with an improvement of $2.49$ \textit{pp}. That is, our method is notably better in those scenarios where we need a higher intersection between the predicted bounding box and the ground truth. 

\begin{table}[h]
\centering
\resizebox{0.86\linewidth}{!}
{
\begin{tabular}{c|cc|cc}
\hline 
\multirow{2}{*}{Method}   & \multicolumn{2}{c|}{SSD 512 (512$\times$512)}     & \multicolumn{2}{c}{SSD 300 (300$\times$300)}  \\
          & IoU\textgreater0.5 & IoU\textgreater0.75 &  IoU\textgreater0.5 & IoU\textgreater0.75                                                                  \\ \hline
SSD~\cite{liu2016ssd}     & 73.22$\pm$0.35   & 45.74$\pm$0.70   & 69.29$\pm$0.51   & 43.36$\pm$1.24     \\
$Ours_{gmm}$    & 73.50$\pm$0.12   & 48.23$\pm$0.53  & 70.19$\pm$0.36  & 46.11$\pm$0.38   \\ 
$Ours_{eff}$    & 73.73$\pm$0.16   & 48.12$\pm$0.33    & 70.45$\pm$0.06     & 46.18$\pm$0.26                                 \\ \hline 
\end{tabular}
}
\caption{\textbf{VOC07:} mAP (in \%) as a function of the resolution of the input image.}
\label{SSD512}
\vspace{-8pt}
\end{table}

\subsection{Accuracy as a function of budget number in active learning}
In the main paper, we used a budget of 1k following the setup of~\cite{DBLP:conf/cvpr/YooK19} to enable direct comparison on VOC07+12. In order to check for the mAP with respect to a budget number in active learning, we further compare the mAP for cases of 9k, 3k, and 1k as the budget number. We summarize the results of this experiment in \tab{table_voc0712_budget}. As in the main paper, we report the performance using the average of mAP and standard deviation for three independent trials. As shown in the last active learning iteration, as expected, we can see that the smaller the budget number yields a higher accuracy improvement in active learning.

\begin{table}[h]
\centering
\resizebox{0.84\linewidth}{!}{
\begin{tabular}{c|c|c|c}
\hline
\multirow{2}{*}{\# of images} & \multicolumn{3}{c}{Budget number} \\ %\cline{2-4}
 & 9k & 3k & 1k \\ \hline
1k    &     0.5254$\pm$0.0017 & 0.5254$\pm$0.0017 & 0.5254$\pm$0.0017\\
2k    &     - & - & 0.6130$\pm$0.0040\\
3k    &     - & - & 0.6556$\pm$0.0051\\
4k    &     - & 0.6797$\pm$0.0011 & 0.6843$\pm$0.0043\\
5k    &     - & - & 0.7077$\pm$0.0019\\
6k    &     - & - & 0.7252$\pm$0.0027\\
7k    &     - & 0.7335$\pm$0.0049 & 0.7352$\pm$0.0025\\
8k    &     - & - & 0.7453$\pm$0.0025\\
9k    &     - & - & 0.7509$\pm$0.0014\\
10k   &     0.7493$\pm$0.0043 & 0.7550$\pm$0.0012 & 0.7598$\pm$0.0021\\ \hline
\end{tabular}}
\caption{\textbf{VOC07+12:} mAP as a function of the budget number in active learning.}
\label{table_voc0712_budget}
\vspace{-6pt}
\end{table}

\begin{figure*}[!]
	\centering
	\begin{tabular}{cc|cc}
	\includegraphics[scale=0.3]{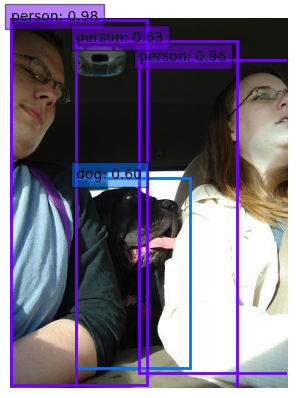}&
    \includegraphics[scale=0.3]{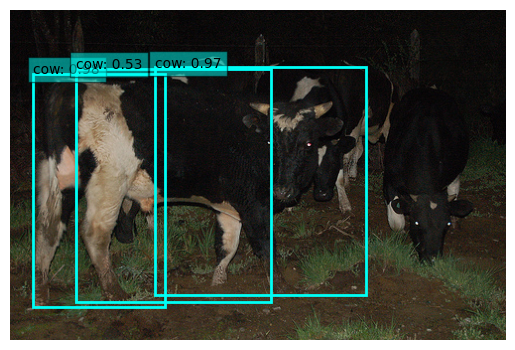}&
	\includegraphics[scale=0.3]{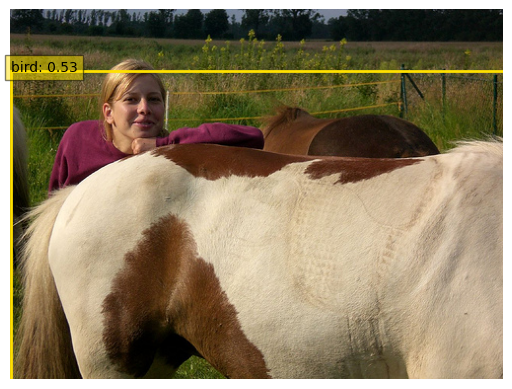}&
	\includegraphics[scale=0.3]{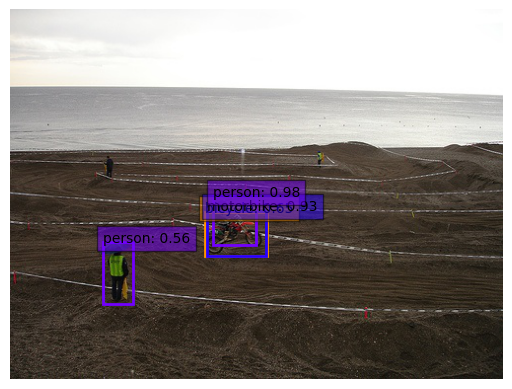}\\
	\scriptsize{\begin{tabular}{ll}$u_{al_b}$: \textbf{3.57} & $u_{al_c}$: $0.57$\\ $u_{ep_b}$: $0.08$ & $u_{ep_c}$: $-0.30$ \end{tabular}}&
	\scriptsize{\begin{tabular}{ll}$u_{al_b}$: $1.49$ & $u_{al_c}$: $0.86$ \\ $u_{ep_b}$: \textbf{10.39}& $u_{ep_c}$: $-0.32$ \end{tabular}}&
	\scriptsize{\begin{tabular}{ll}$u_{al_b}$: $-0.82$ & $u_{al_c}$: \textbf{11.71}\\ $u_{ep_b}$: $-0.37$ & $u_{ep_c}$: $1.31$  \end{tabular}}&
	\scriptsize{\begin{tabular}{ll}$u_{al_b}$: $1.47$ & $u_{al_c}$: $0.63$\\ $u_{ep_b}$: $0.05$ & $u_{ep_c}$: \textbf{11.84} \end{tabular}}%\\	(a)&(b)&(c)&(d)
	\\
	\includegraphics[scale=0.3]{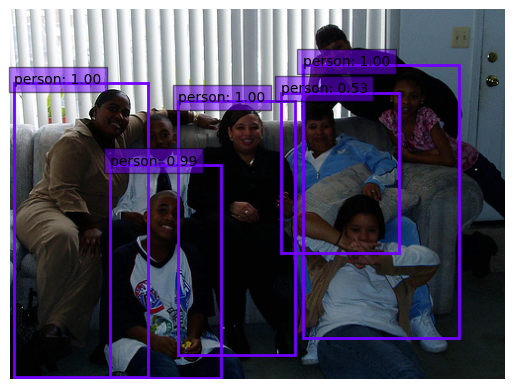}&
	\includegraphics[scale=0.3]{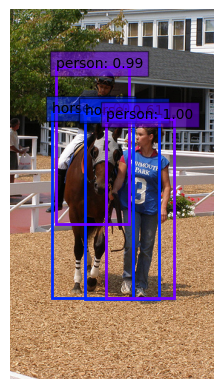}&
	\includegraphics[scale=0.3]{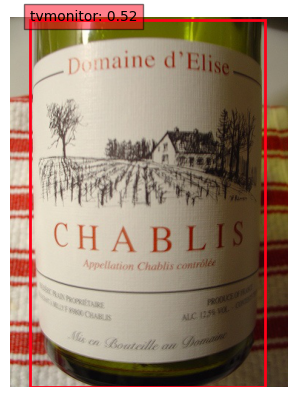}&
	\includegraphics[scale=0.3]{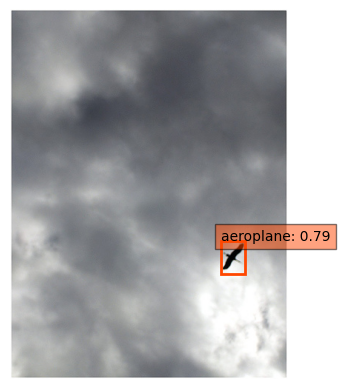}\\
	\scriptsize{\begin{tabular}{ll}$u_{al_b}$: \textbf{3.45} & $u_{al_c}$: $-0.57$\\ $u_{ep_b}$: $0.88$ & $u_{ep_c}$: $0.05$ \end{tabular}}&
    \scriptsize{\begin{tabular}{ll}$u_{al_b}$: $1.56$ & $u_{al_c}$: $-0.17$ \\ $u_{ep_b}$: \textbf{9.54}& $u_{ep_c}$: $0.47$ \end{tabular}}&
	\scriptsize{\begin{tabular}{ll}$u_{al_b}$: $-0.61$ & $u_{al_c}$: \textbf{8.30}\\ $u_{ep_b}$: $-0.38$ & $u_{ep_c}$: $0.62$  \end{tabular}}&
	\scriptsize{\begin{tabular}{ll}$u_{al_b}$: $-0.19$ & $u_{al_c}$: $1.58$\\ $u_{ep_b}$: $-0.35$ & $u_{ep_c}$: \textbf{10.06}\end{tabular}}\\
	\includegraphics[scale=0.3]{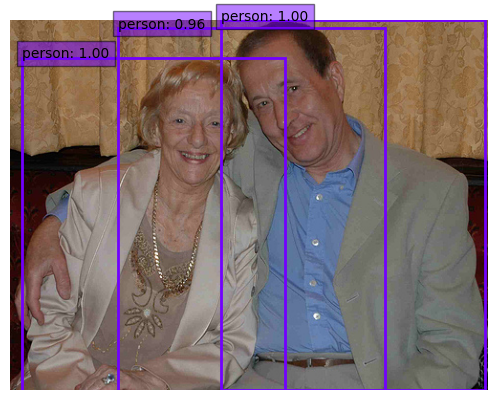}&
	\includegraphics[scale=0.3]{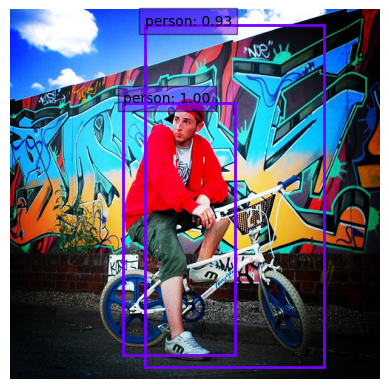}&
	\includegraphics[scale=0.3]{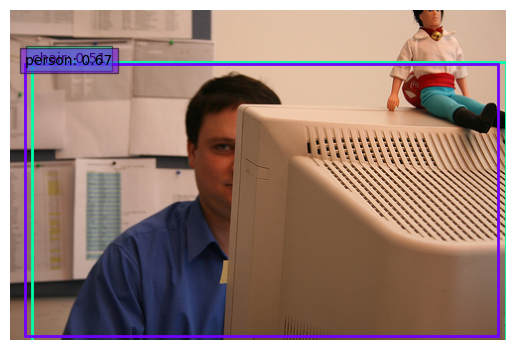}&
	\includegraphics[scale=0.3]{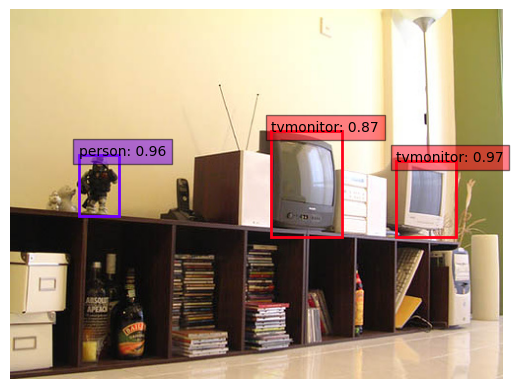}\\
	\scriptsize{\begin{tabular}{ll}$u_{al_b}$: \textbf{3.59} & $u_{al_c}$: $-0.37$\\ $u_{ep_b}$: $-0.30$ & $u_{ep_c}$: $-0.39$ \end{tabular}}&
	\scriptsize{\begin{tabular}{ll}$u_{al_b}$: $1.77$ & $u_{al_c}$: $-0.75$ \\ $u_{ep_b}$: \textbf{8.10}& $u_{ep_c}$: $0.06$ \end{tabular}}&
	\scriptsize{\begin{tabular}{ll}$u_{al_b}$: $-0.39$ & $u_{al_c}$: \textbf{7.49}\\ $u_{ep_b}$: $-0.36$ & $u_{ep_c}$: $1.19$  \end{tabular}}&
	\scriptsize{\begin{tabular}{ll}$u_{al_b}$: $0.47$ & $u_{al_c}$: $1.19$\\ $u_{ep_b}$: $-0.12$ & $u_{ep_c}$: \textbf{6.92} \end{tabular}}\\
	\end{tabular}
	%\vspace{-.15cm}
	\caption{Examples of aleatoric and epistemic uncertainties for inaccurate detections. Best viewed in screen.}
	\label{fig:img_examples_more}
	%\vspace{-.5cm}
\end{figure*}

\section{More visual examples selected by our approach}
\fig{fig:img_examples_more} shows more representative examples selected by our active learning approach. Each uncertainty value (bold numbers in~\fig{fig:img_examples_more}) provides a different insight into some particular failure. From left to right and top to bottom: One of the several bounding boxes detected as person is false positive; One of the several bounding boxes detected as cow is false positive; A horse is misclassified as a bird; A motorbike is misclassified as bicycle; One of the several bounding boxes detected as person is false positive; One of the several bounding boxes detected as horse is false positive; A bottle is misclassified as a TV/monitor; A bird is misclassified as an aeroplane; One of the several bounding boxes detected as person is false positive; One of the several bounding boxes detected as person is false positive; A person is misclassified as a chair; A toy (not in the PASCAL VOC dataset) is misclassified as a person.

\begin{table*}[!]
\centering
\resizebox{0.87\linewidth}{!}{
\begin{tabular}{c|cccc|cc}
\hline 
\# of labeled images & Random~\cite{liu2016ssd}        & Entropy~\cite{DBLP:conf/bmvc/RoyUN18}        & Core-set~\cite{DBLP:conf/iclr/SenerS18}      & LLAL~\cite{DBLP:conf/cvpr/YooK19}          & $Ours_{gmm}$    & $Ours_{eff}$ \\ \hline
1k    & 0.5262$\pm$0.0062 & 0.5262$\pm$0.0062 & 0.5262$\pm$0.0062 & 0.5238$\pm$0.0028 &  0.5254$\pm$0.0017 & 0.5254$\pm$0.0017\\
2k    & 0.6082$\pm$0.0019 & 0.6123$\pm$0.0081 & 0.6236$\pm$0.0052 & 0.6095$\pm$0.0042 &  0.6130$\pm$0.0040 & 0.6121$\pm$0.0050\\
3k    & 0.6423$\pm$0.0022 & 0.6357$\pm$0.0091 & 0.6590$\pm$0.0043 & 0.6491$\pm$0.0047 &  0.6556$\pm$0.0051 & 0.6657$\pm$0.0027\\
4k    & 0.6633$\pm$0.0018 & 0.6694$\pm$0.0021 & 0.6763$\pm$0.0021 & 0.6690$\pm$0.0028 &  0.6843$\pm$0.0043 & 0.6849$\pm$0.0014\\
5k    & 0.6751$\pm$0.0017 & 0.6870$\pm$0.0015 & 0.6888$\pm$0.0048 & 0.6905$\pm$0.0045 & 0.7077$\pm$0.0019 & 0.7073$\pm$0.0012\\
6k    & 0.6860$\pm$0.0050 & 0.6982$\pm$0.0011 & 0.6944$\pm$0.0032 & 0.7035$\pm$0.0055 & 0.7252$\pm$0.0027 & 0.7185$\pm$0.0016\\
7k    & 0.6927$\pm$0.0016 & 0.7018$\pm$0.0027 & 0.7016$\pm$0.0013 & 0.7149$\pm$0.0066 & 0.7352$\pm$0.0025 & 0.7318$\pm$0.0045\\
8k    & 0.7010$\pm$0.0017 & 0.7112$\pm$0.0012 & 0.7083$\pm$0.0012 & 0.7213$\pm$0.0060 & 0.7453$\pm$0.0025 & 0.7429$\pm$0.0044\\
9k    & 0.7044$\pm$0.0047 & 0.7166$\pm$0.0031 & 0.7115$\pm$0.0016 & 0.7273$\pm$0.0030 & 0.7509$\pm$0.0014 & 0.7483$\pm$0.0028\\
10k   & 0.7117$\pm$0.0016 & 0.7222$\pm$0.0024 & 0.7171$\pm$0.0025 & 0.7338$\pm$0.0028 & 0.7598$\pm$0.0021 & 0.7584$\pm$0.0026\\ \hline 
\end{tabular}
}
\caption{\textbf{VOC07+12:} Comparison to published work using a single model for scoring.}
\label{table_voc0712_single}
\end{table*}

\begin{table*}[!] 
\centering
\resizebox{0.88\linewidth}{!}{
\begin{tabular}{c|ccc|cc}
\hline 
\# of labeled images & MC-dropout~\cite{feng2019deep} (50 fwd) & MC-dropout~\cite{feng2019deep} (25 fwd)    & Ensemble~\cite{haussmann2020scalable}      & $Ours_{gmm}$    & $Ours_{eff}$ \\ \hline
1k    &     0.5235 $\pm$ 0.0004  & 0.5235$\pm$0.0004 & 0.5254$\pm$0.0017 & 0.5254$\pm$0.0017 & 0.5254$\pm$0.0017\\
2k    &     0.6059 $\pm$ 0.0026  & 0.6059$\pm$0.0028 & 0.6020$\pm$0.0093 & 0.6130$\pm$0.0040 & 0.6121$\pm$0.0050\\
3k    &     0.6660 $\pm$ 0.0023  & 0.6690$\pm$0.0030 & 0.6570$\pm$0.0099 & 0.6556$\pm$0.0051 & 0.6657$\pm$0.0027\\
4k    &     0.6890 $\pm$ 0.0018  & 0.6840$\pm$0.0019 & 0.6920$\pm$0.0034 & 0.6843$\pm$0.0043 & 0.6849$\pm$0.0014\\
5k    &     0.7060 $\pm$ 0.0045  & 0.7080$\pm$0.0041 & 0.7150$\pm$0.0018 & 0.7077$\pm$0.0019 & 0.7073$\pm$0.0012\\
6k    &     0.7200 $\pm$ 0.0012  & 0.7190$\pm$0.0050 & 0.7290$\pm$0.0027 & 0.7252$\pm$0.0027 & 0.7185$\pm$0.0016\\
7k    &     0.7367 $\pm$ 0.0015  & 0.7381$\pm$0.0003 & 0.7429$\pm$0.0004 & 0.7352$\pm$0.0025 & 0.7318$\pm$0.0045\\
8k    &     0.7468 $\pm$ 0.0027  & 0.7475$\pm$0.0056 & 0.7491$\pm$0.0041 & 0.7453$\pm$0.0025 & 0.7429$\pm$0.0044\\
9k    &     0.7549 $\pm$ 0.0013  & 0.7558$\pm$0.0023 & 0.7589$\pm$0.0025 & 0.7509$\pm$0.0014 & 0.7483$\pm$0.0028\\
10k   &     0.7567 $\pm$ 0.0048  & 0.7601$\pm$0.0018 & 0.7590$\pm$0.0032 & 0.7598$\pm$0.0021 & 0.7584$\pm$0.0026\\ \hline 
\end{tabular}
}
\caption{\textbf{VOC07+12:} Accuracy comparison to MC-dropout and ensemble. For MC-dropout, we include two instances: using 25 forward passes and using 50 forward passes.}
\label{table_voc0712_multi}
\end{table*}

\begin{table*}[!]
\centering
\resizebox{0.65\linewidth}{!}{
\centering
\begin{tabular}{c|ccc|cc}
\hline 
                     & SSD~\cite{liu2016ssd}  & Ensemble~\cite{haussmann2020scalable} & MC-dropout~\cite{feng2019deep} & $Ours_{gmm}$ & $Ours_{eff}$\\ \hline
\# of parameters (M) & 26.29 & 78.87     & 26.29       & 52.35 &   41.12 \\
Forward time (sec)     & 0.023 & 0.069     & 0.412       & 0.031 &   0.029 \\ 
\hline 
\end{tabular}
}
\caption{\textbf{VOC07+12:} Model parameters in millions (M) and forward time in seconds (sec) using a resolution of $300\times300$ for the input image and $K=4$.}
\label{parameter_time}
\end{table*}

\section{Values in the plots of VOC07+12}
In the main paper, we present plots for active learning results using VOC07+12 in Fig. 4 and Fig. 5. \tab{table_voc0712_single},~\tab{table_voc0712_multi}, and~\tab{parameter_time} summarize the actual numbers used to create the plots. As mentioned in the paper, in~\tab{table_voc0712_single}, numbers corresponding to Random~\cite{liu2016ssd} , Entropy~\cite{DBLP:conf/bmvc/RoyUN18}, Core-set~\cite{DBLP:conf/iclr/SenerS18}, and LLAL~\cite{DBLP:conf/cvpr/YooK19} are taken from~\cite{DBLP:conf/cvpr/YooK19}. For MC-dropout~\cite{feng2019deep}, to further verify the influence in the number of forward passes, we include two instances: using 25 (the one included in the main paper) and 50 forward passes. As we can see in~\tab{table_voc0712_multi}, the variation in accuracy for these two approaches is negligible while the compute needed is significantly larger for the one using 50 forward passes.

\section{Discussion of the classification loss}
In addition to Eq. 5 and Eq. 9 in the main paper (called \textit{Type-1} loss), we can train the proposed object detection network with the following classification loss:
\begin{equation}\small
\label{eq:cls_loss_ver_2}
\begin{gathered}
L^{Pos}_{cl}(y,c)=-\sum_{i\in{Pos}}^Ny_G^{i}~log\sum_{k=1}^K\pi^{ik}Softmax(\hat{c}^{ik}_p)\\
L^{Neg}_{cl}(y,c)=-\sum_{i\in{Neg}}^{M\times{N}}y_0^{i}~log\sum_{k=1}^K\pi^{ik}Softmax(\hat{c}^{ik}_p),
\end{gathered}
\end{equation}
\noindent where $y_G$ and $y_0$ are one-hot vectors having 1 in ground-truth class $G$ and background class $0$, respectively. The remaining parameters are the same as Eq. 5 in the main paper. The parameters of the loss for $Ours_{eff}$ are the same as Eq.~\ref{eq:cls_loss_ver_2} except for the class probability $\hat{\mu}^{ik}_p$. For \textit{Type-1} loss, the weights tend to be concentrated in one of the mixture distributions in training. For Eq.~\ref{eq:cls_loss_ver_2} (called \textit{Type-2} loss), however, this trend tends to be alleviated. \tab{other_cls_loss_voc} and~\tab{other_cls_loss_coco} show the results of active learning of two classification losses on VOC07 and MS-COCO, respectively. As shown, there is no significant difference in the accuracy of the two loss functions on VOC07, but there is a large difference on MS-COCO. Although the \textit{Type-2} loss mitigates the weight bias compared to the \textit{Type-1} loss, the accuracy improvement is not sufficient for larger dataset with more classes. For this reason, in the main paper, we show the experimental results based on the \textit{Type-1} loss, but a study on a classification loss design that can improve the overall active learning performance while resolving the weight bias is needed in the future.

\begin{table}[h]
\centering
\resizebox{0.86\linewidth}{!}{
\centering
\begin{tabular}{c|c|ccc}
\hline 
\multirow{2}{*}{Model} & \multirow{2}{*}{Cls. loss} & \multicolumn{3}{c}{mAP in \% (\# images)} \\
                    &  & 1st (2k)     & 2nd (3k)    & 3rd (4k)     \\\hline
\multirow{2}{*}{$Ours_{gmm}$} & Type-1       & 62.43$\pm$0.10      & 67.32$\pm$0.12      & 69.43$\pm$0.11 \\
& Type-2       & 62.64$\pm$0.21      & 67.20$\pm$0.22      & 69.40$\pm$0.14    \\ \hline
\multirow{2}{*}{$Ours_{eff}$} & Type-1     & 62.91$\pm$0.16      & 67.61$\pm$0.17      & 69.66$\pm$0.17   \\
& Type-2       & 62.23$\pm$0.25      & 67.30$\pm$0.23      & 69.57$\pm$0.16     \\ \hline
\end{tabular}}
\caption{Ablation study of classification losses on \textbf{VOC07}.}
\label{other_cls_loss_voc}
%\vspace{-6pt}
\end{table}

\begin{table}[h]
\centering
\resizebox{0.86\linewidth}{!}{
\centering
\begin{tabular}{c|c|ccc}
\hline 
\multirow{2}{*}{Model} & \multirow{2}{*}{Cls. loss} & \multicolumn{3}{c}{mAP in \% (\# images)} \\
                    &  & 1st (5k)     & 2nd (6k)    & 3rd (7k)     \\\hline
\multirow{2}{*}{$Ours_{gmm}$} & Type-1       & 27.70$\pm$0.08    & 29.28$\pm$0.05     & 30.51$\pm$0.12  \\
& Type-2       & 27.38$\pm$0.16             & 28.69$\pm$0.22      & 29.55$\pm$0.14    \\ \hline
\multirow{2}{*}{$Ours_{eff}$} & Type-1     & 27.33$\pm$0.04    & 29.06$\pm$0.08     & 30.02$\pm$0.05    \\
& Type-2       & 27.46$\pm$0.13      & 28.32$\pm$0.32      & 29.21$\pm$0.14     \\ \hline
\end{tabular}}
\caption{Ablation study of classification losses on \textbf{MS-COCO}.}
\label{other_cls_loss_coco}
\vspace{-0.3cm}
\end{table}

% %\newpage
% {\small
% \bibliographystyle{ieee_fullname}
% \bibliography{egbib}
% }

%\newpage
{\small
\bibliographystyle{ieee_fullname}
\bibliography{egbib}
}

\end{document}